\documentclass[conference]{IEEEtran}
\usepackage{times}
\usepackage{epsfig}
\usepackage{graphicx}
\usepackage{amsmath}
\usepackage{amssymb}
\usepackage{color}
\usepackage{booktabs}
\usepackage{multirow}
\usepackage{algorithm}
\usepackage{algorithmic}
\usepackage{caption}
\usepackage{subcaption}

\usepackage[breaklinks=true,bookmarks=false]{hyperref}

\def\BibTeX{{\rm B\kern-.05em{\sc i\kern-.025em b}\kern-.08em
    T\kern-.1667em\lower.7ex\hbox{E}\kern-.125emX}}
    
\begin{document}



\title{ A Novel Design of Adaptive and Hierarchical Convolutional Neural Networks using Partial Reconfiguration on FPGA}

\author{Mohammad Farhadi\\
Arizona State University\\
Tempe, AZ, USA\\
{\tt\small mfarhadi@asu.edu}
\and
Mehdi Ghasemi\\
Arizona State University\\
Tempe, AZ, USA\\
{\tt\small mghasem1@asu.edu}
\and
Yezhou Yang\\
Arizona State University\\
Tempe, AZ, USA\\
{\tt\small yz.yang@asu.edu}
}

\maketitle
%
%
%
%

 
\begin{abstract}
Nowadays most research in visual recognition using Convolutional Neural Networks (CNNs) follows the ``deeper model with deeper confidence'' belief to gain a higher recognition accuracy. At the same time, deeper model brings heavier computation. On the other hand, for a large chunk of recognition challenges, a system can classify images correctly using simple models or so-called shallow networks. Moreover, the implementation of CNNs faces with the size, weight, and energy constraints on the embedded devices. In this paper, we implement the adaptive switching between shallow and deep networks to reach the highest throughput on a resource-constrained MPSoC with CPU and FPGA. 
To this end, we develop and present a novel architecture for the CNNs where a gate makes the decision whether using the deeper model is beneficial or not. Due to resource limitation on FPGA, the idea of partial reconfiguration has been used to accommodate deep CNNs on the FPGA resources. We report experimental results on CIFAR-10, CIFAR-100, and SVHN datasets to validate our approach. Using confidence metric as the decision making factor, only 69.8\%, 71.8\%, and 43.8\% of the computation in the deepest network is done for CIFAR-10, CIFAR-100, and SVHN while it can maintain the desired accuracy with the throughput of around 400 images per second for SVHN dataset. \hyperlink{https://github.com/mfarhadi/AHCNN}{https://github.com/mfarhadi/AHCNN}.    
%
%
\end{abstract}

\section{Introduction}
\label{sec:intro}



Recently, Convolutional Neural Networks (CNNs)-based methods achieve great success in image classification \cite{Krizhevsky2012} and object detection \cite{redmon2016yolo9000} tasks. The success leads the researchers to explore deeper models such as ResNet\cite{He2016} (152 layers), and these models yield high recognition accuracy. The ``secret'' sauce of success for these deeper and deeper CNNs models are stacking repetitive layers and increasing the number of model parameters. This practice is possible while the applications are running in big data centers or infrastructures with high performance processing capabilities. However, these complex models are not suitable for real-time and embedded systems due to low energy constraints and limited computing resources.  



The aforementioned concern triggers various approaches, such as by the alignment of memory and SIMD (Single instruction, multiple data) operations to boost matrix operations (93\% Top-5 accuracy) \cite{gong2014compressing}, specific hardware (FPGA) solutions (86.66\% Top-5 accuracy) \cite{qiu2016going}, network compression (89.10\% Top-5 accuracy) \cite{han2015deep} or using cloud computing (network latency should be considered) \cite{Chen2015}. These approaches indeed can reduce the energy consumption, but they fail to retain recognition accuracy while a system faces critical situations. In other words, they reduce the computation overload by trading a large chunk of recognition accuracy off from the state-of-the-art performances which is more than 96\% Top-5 accuracy at the moment. 


\begin{figure}[]
    \centering
    \includegraphics[trim=2.3cm 3.8cm 10cm 3.0cm, clip=true, width=6cm]{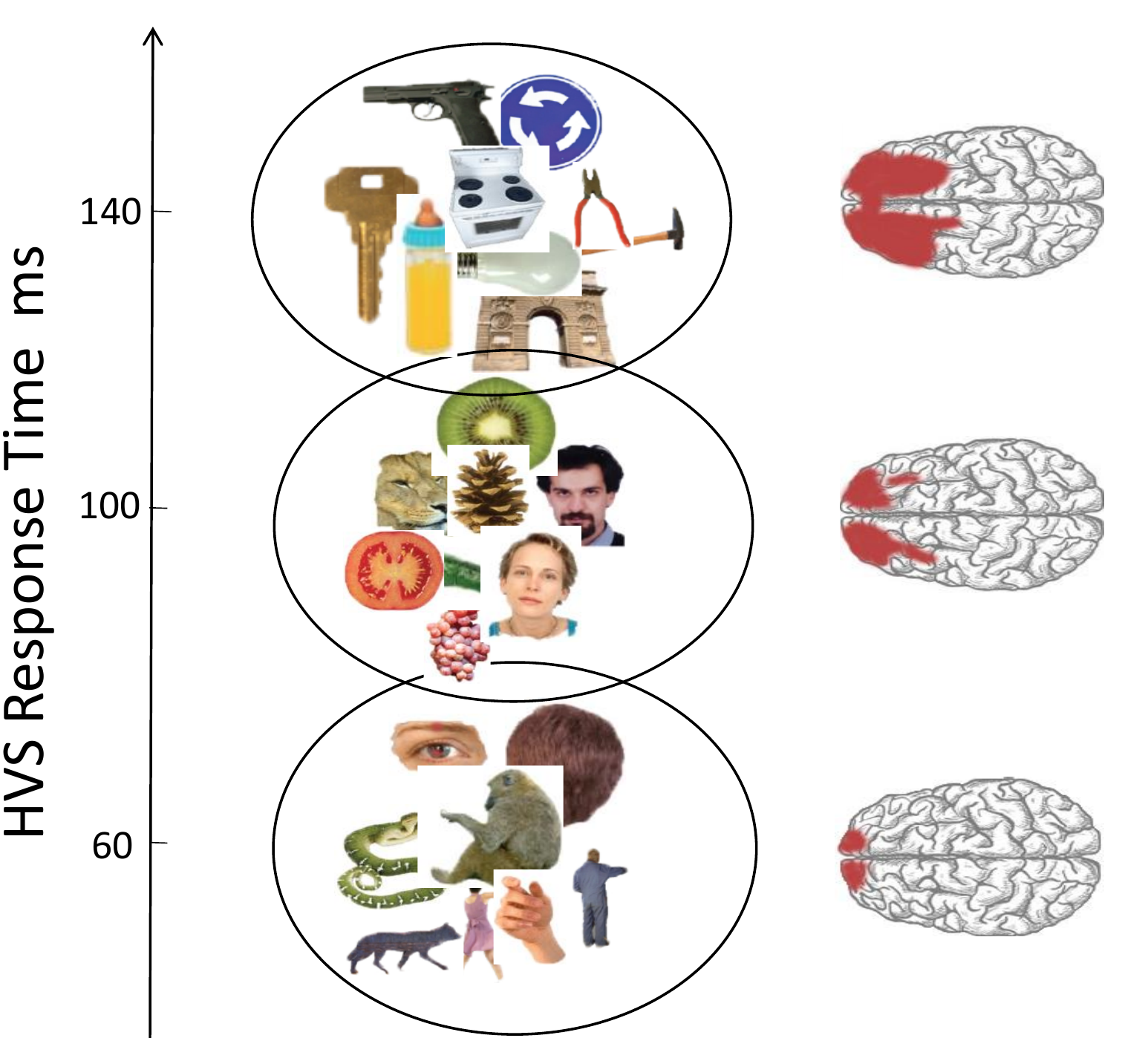}
    \caption{Stimulus materials, fMRI brain coverage, and significant MEG-fMRI fusion results over different objects and background \cite{cichy2016similarity}.}
    \label{brain-model}
\end{figure}

On the other hand, recent research suggests that the human vision system (HVS) has two stages for conducting visual classification: 1) a shallow primary stage and 2) a decision layer to pick a further processing pathway \cite{Ritchie2015}. The study also supports the theory that the structure of the object representation in the HVS influences the decision layer during visual classification. Results from another research that is conducted in the field of neuroscience  \cite{Cichy2014} showed that the response time of our HVS given an specific image as the stimuli differs a lot based on the category that the image belongs to. These results again suggest that HVS has a decision system which controls our processing resources assigned for each image. The study \cite{cichy2016similarity} shows the brain behavior corresponding to the images from different  categories (in Fig.~\ref{brain-model}).  From the fMRI imaging, researchers have speculation  that for some input images, only a ``shallow'' part of our HVS is utilized while for other categories they invoke a ``deeper'' processing.


Following these insights and observations, we design and implement a feedback procedure to determine the early exit from the model. The core part of this procedure is an engine that accesses an image and predicts how accurate a certain model will perform. Our proposed model, besides classifying images, has an extra output which is designed to provide an evaluation on how well a model will perform. Then, our system relies on this evaluation to decide whether classifying this image with the deeper model will be beneficial or not. We implement a gate operation which takes the evaluation and it has an adjustable tolerance threshold for decision making. For example, in autonomous driving scenario, if the class ``human'' appears in the top 5 or 10 results from the shallow model, the system can adaptively decrease the decision layer threshold and yields a more accurate prediction using deeper models. Thus, our feedback procedure optimizes the resource usage by controlling the type and amount of images being sent to the deeper model.     


In this paper, we report the implementation of the feedback procedure for determining the path of inference in the CNN based on the confidence level factor. The method has been implemented on an MPSoC (Pynq-Z1) with an ARM CPU and FPGA. We propose and implement the idea of partial reconfiguration in the FPGA to map the quantized CNN on the FPGA resources. We further demonstrate it using ResNet CNNs \cite{He2016} on CIFAR-10, CIFAR-100, and SVHN datasets. The experimental results show that on average only 69.8\%, 71.8\%, and 43.8\% of computation on the deepest network needs to be utilized  for CIFAR-10, CIFAR-100, and SVHN benchmarking datasets, to maintain a comparablely high recognition performance. 



		

\section{Related Work}


The last few years in the field of Deep Learning has laid the foundation for major advancements in visual recognition systems, ranging from object recognition~\cite{Krizhevsky2012,he2016deep}, action recognition~\cite{lea2016learning}, to  scene recognition~\cite{zhou2014learning}. Significant improvements in recognition accuracies allow a wide range of science fiction ideas materialized, resulting in economic and societal benefits with AI applications such as autonomous vehicles \cite{chen2015deepdriving}, intelligent IoT systems \cite{tang2017enabling}, industrial robots, service robots and intelligent health care systems \cite{ravi2017deep,izadyyazdanabadi2017convolutional}.

The increasing number of real world applications require their corresponding visual recognition engine to not only recognize well, but also actively and effectively adjust its computational resources to handle the ever-changing physical world situations that the systems will face. The seminal work of the cascaded classification of Viola and Jones \cite{viola2001rapid} represents the line of studies on cost-sensitive classification. The essence of their work is to treat classification as a cascaded process that contains control layers deciding the exit points where the system is confident in its current inference. Following the similar line of work, more recently,  \cite{li2015convolutional, shen2017fast} have proposed cascading CNNs structures to reduce the computational cost by reducing the structural complexities of CNNs. 

Another thrust of work has focused on reducing the resource consumption of CNNs or other types of neural networks through various techniques of compressing the network structures \cite{ba2014deep,han2016deep,han2015learning,rastegari2016xnor}. Network pruning is one of the well-studied approaches which removes unnecessary nodes and edges from network, to compress model and gaining inference speedup \cite{han2016deep,wen2016learning,iandola2016squeezenet}.  However, \cite{han2015learning} pointed out that using the standard GPU implementation, the speedup is hard to achieve due to the lack of high degrees of exploitable regularity and computation intensity in the resulting network with sparse connections. 

The use of adaptive structures is a relatively newer approach which decides how to further process the image  \cite{shen2017fast,zhou2017adaptive,teerapittayanon2016branchynet,bengio2015conditional}. Teerapittayanon et al \cite{teerapittayanon2016branchynet} proposed an adaptive model to allow early exit based on the entropy of model output which is called Branchy-Net. By adding sub-outputs to the model, Branchy-Net checks the entropy of model output and if the entropy is low enough, terminates the procedure. By doing this, Branchy-Net achieved 2x speedups at the inference time \cite{teerapittayanon2016branchynet}. However, Branchy-Net spends a considerable amount of time to evaluate the early output \cite{bolukbasi2017adaptive}; it does not a have clear procedure to select the location of early branches, and it is changing the structure of original model to have an early exit. In response to the mentioned issue, Bollukbasi et al \cite{bolukbasi2017adaptive} proposed an adaptive method which adopts the Branchy-Net idea and stacks several models such as AlexNet \cite{Krizhevsky2012} and ResNet \cite{he2016deep}. This model still suffers from the overhead time of evaluating the model's early output. Another study in \cite{wang2018skipnet}, proposed a method based on a decision gate to skip some of the blocks in ResNet structure. The decision gates include convolution and fully connected layers which are trained using reinforcement learning. These decision gates are not suitable for shallow CNN models such as ResNet-18 \cite{He2016}.    

The implementation of CNNs on FPGAs has been  studied from the literature to certain extent. More specifically, \textit{BinaryEye} in \cite{jokic2018binaryeye} has presented an implementation of binary neural networks on FPGA. The presented implementation can be used in IoT and distributed systems where the stream of images for a camera needs to be processed. A framework called \textit{FINN} has been also presented in \cite{umuroglu2017finn} for the inference of binarized neural networks. The mentioned implementation does not adopt the partial reconfiguration to address the limitation of resources on FPGA.     

Dynamic partial reconfiguration has been done in the relevant literature in \cite{al2013dynamic,kastner2018hardware}. In \cite{al2013dynamic}, the authors have implemented the reconfiguration steps in a Zynq 7000 FPGA but do not implement CNN architectures. Dynamic reconfiguration has been done in \cite{kastner2018hardware} for the CNNs on the Pynq board. In the mentioned work, they have stated that the implementation of CNN using reconfiguration at each layer is expensive.

In this paper, we have implemented the idea of adaptive switching between shallow and deep networks on FPGA platform using partial reconfiguration to reduce the amount of needed computation. The confidence level was observed to be the most efficient factor to switch in comparison with the methods presented in Skip-Net \cite{wang2018skipnet} and Branchy-Net \cite{teerapittayanon2016branchynet}.

\section{Adaptive and Hierarchical CNNs}
\label{sec:method}
The key module of our proposed  Adaptive and Hierarchical convolutional neural networks (AH-CNN) model is a feedback procedure which is designed to comprehensively evaluate the classification procedure. More specifically, AH-CNN consists of three parts: 1) a \textbf{shallow part} which is a light-weight CNN model; 2) a \textbf{decision layer} which evaluates shallow part's performance and makes a decision; and 3) a \textbf{deep part} which is a deep CNN  with a high inference accuracy.  As mentioned in Section~\ref{sec:intro}, the overall objective of our dynamic system is to obtain the highest possible recognition accuracy during critical time instances while  maintaining a satisfiable performance using the shallow part during non-critical moments. Following this intuition, we put forward a mechanism with a combination of a shallow model, feedback procedure and a deep model, which has a flexible structure at the same time.  This mechanism can achieve the same high recognition accuracy as other very deep networks by partially reconfiguring the hardware structure. Thus, an intelligent agent equipped with the AH-CNN can adaptively adjust its model structure to maintain a balance between the expected classification accuracy and the model complexity. This procedure can be applied repetitively and has several decision layers. In the following section, we will describe the details of the AH-CNN architecture.



\subsection{AH-CNN Architecture}

The authors in \cite{Zeiler2014} showed that the preceding layers in deep neural networks respond to class-agnostic low-level features, while the rear layers extract more specific high-level features. Objects of certain categories can be classified solely by the low-level features but for the images of other categories, we need more specific high-level features, and deeper layers are needed to extract them. Thus, we design  our architecture to have three modules: the shallow part, the deep part and a decision layer. Hence, the proposed AH-CNN with a design of an adaptive and hierarchical structure, can yield different behaviors based on the input image characteristics. We will describe the three mentioned modules in the following.

\textbf{Shallow Part:} In this work, the FPGA is loaded with the shallow part first. This part can be applied to the input tensor without any reconfiguration cost and classifies all input images and it outputs two results: 1) a predicted label $y=j$ and 2) a confidence value   ($P(y=j|X_{i})=softmax(z_{j})=exp(z_{j})/\sum_{k}^{}exp(z_{k})$, where $z$ is the output of fully connected layer over the input image $X_{i}$) which will be later used in the feedback procedure. 

\textbf{Deep Part:} This part is the next group of convolution layers which should be loaded on FPGA. Due to the transfer and configuration time, loading the new part on the FPGA is expensive. This group of convolution layers is responsible to extract more specific high-level features and detect the images which are misclassified by the shallow part. This part will be applied over the output of the last convolution in the shallow part to reach higher confidence.   

\textbf{Decision Layer:} This part of AH-CNN takes the shallow part's outputs and makes a decision to whether activate the deep part, or simply terminate further processing and take the shallow part's result as the overall model output.  This layer has a feedback procedure to make the network behavior decision by evaluating the shallow part. 



To this end, the decision layer currently yields a binary behavior based on three factors: 1) the confidence value from the shallow part; 2) the priority of the object classes; 3) the overall expected classification accuracy (which is obtained by validating the model over the data set). The binary behavior either activates the deep part or takes the shallow part's classification output as the overall model's output. 

\begin{algorithm}[!ht]
\caption{AH-CNN: Inference Phase}
\begin{algorithmic} 
\footnotesize
\REQUIRE Input image $X_{i}$ , Desired accuracy $\Lambda$, Number of early branches $N_{i}$, High priority classes $S_{HP}$. 
\WHILE{$X_{i}$}
\WHILE{$N_{i}$}
\STATE Assign proper $\Gamma$ based on $\Lambda$
\STATE $ \beta , ShOutput \gets ForwardPropagate(X_{i}, Shallow) $
\IF{ $S_{HP}$ appear in ShOutput Top-n}
\STATE $\Gamma = \Gamma + \Theta $
\ENDIF
\IF{$\beta <= \Gamma$}
\STATE Load deep part on FPGA
\STATE $Output \gets ForwardPropagate((ShOutput,Deep)$
\ELSE
\STATE $ Output \gets ShOutput$
\ENDIF
\ENDWHILE
\ENDWHILE
\end{algorithmic}
\label{testing-phase}
\end{algorithm}
\begin{figure*}[ht]
	\begin{center}
		
		\includegraphics[trim=0.5cm 6.4cm 0.5cm 2.6cm, clip=true,scale=0.50]{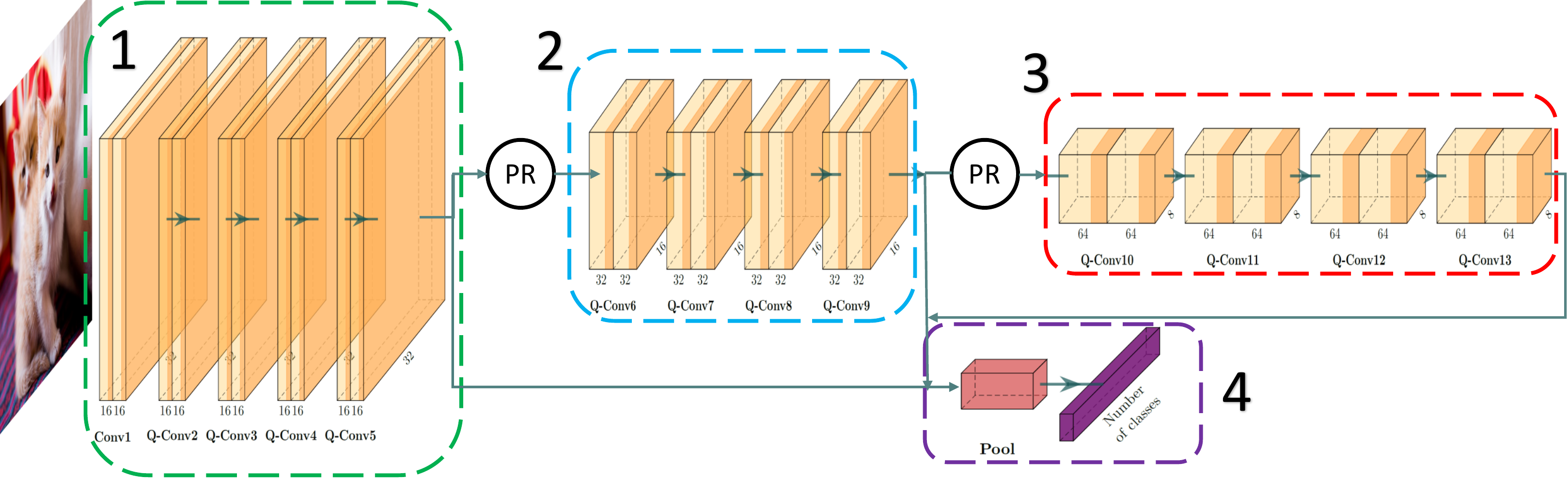}
	\end{center}
	\caption{The scheme of CNN implementation on FPGA using dynamic reconfiguration and adaptive feedback. The adaptive feedback makes the decision to classify the image or apply the next stack of convolution layers based on the output confidences computed from each part.}
	\label{fig:CNNFPGA}
\end{figure*}
Algorithm~\ref{testing-phase} shows the AH-CNN processing procedure in the inference phase. The decision layer first checks the top-$n$ classification results  from the shallow part's classification vector. If a label from the high priority set ($S_{HP}$) exists, there is a higher probability that the input needs further processing. Next, the decision layer checks the current expected classification accuracy, which will affect the fraction of all the input images that need further processing. Finally, the model checks the shallow part's Confidence value. The interpretation of the confidence value yields a feedback procedure. The priority of the object classes and the overall expected classification accuracy  are then considered to tune a threshold value to compare with the confidence value,  which we refer to it as the trigger point later. 

The most critical element of the feedback procedure of AH-CNN is the trigger point $\Gamma$. After feed-forwarding each image over the shallow part, the decision layer gets the confidence value $\beta$ and compares it with the assigned threshold $\Gamma$. If $\beta$ does not reach $\Gamma$, it means that the shallow part has less confidence than our system's tolerance over the input image and further processing is needed to gain a higher expected accuracy. As a consequence, the decision layer load and activates the deep part. The value of the trigger point can be actively adapted according to the real-world situations. In cases that we do not need a high accuracy,  we can decrease the trigger point value.  In cases that the member of $S_{HP}$ appear in the top-$n$ outputs,  we can increase the trigger value ($\Gamma$) by $\Theta$ to expect a higher classification accuracy over that image. The trigger point makes our model innately adaptive. We discuss how to set a proper trigger point as well as its range in Section.~\ref{sec:training}.

\section{Implementation on FPGA}
The overall scheme of implementation on the FPGA is depicted in Figure \ref{fig:CNNFPGA}. The convolution layers in the CNN are based on the ResNet CNN structure. The whole CNN is divided into three parts which are numbered in the figure. The output of each part can be used as the input for the pooling layer in part 4. There is a Partial Reconfiguration unit labeled as \textit{PR} which changes the bitstream file on the FPGA when necessary. The reason for partial reconfiguration is to save the LUT area on the FPGA and address the limitation of computational resources. 

In order to implement the CNN on FPGA, a quantized version of CNN has been used which is popular in the FPGA community \cite{umuroglu2017finn}. In this network, the weights are binary and the activation data are five bits (quantized bits). Even using this quantization method and binary values, an acceptable accuracy of classification can be obtained which is shown in the experimental result section. 

Batch processing has been used to improve the overall throughput of the system. During batch processing, the reconfiguration overhead of changing the bitstream files would be considered for all the images that are going to be processed in the network. Therefore, the overhead of reconfiguration would be negligible when calculating the inference time for one image on average.

\section{Training Phase}
Both the shallow and the deep part aim to classify images with the best possible performance that can be achieved individually. Consequently, the feedback procedure should not have any influence over the shallow part's classification performance. We train both the deep part and the shallow part using the stochastic gradient mini-batch \cite{dean2012large}. 
Also, the mean and range of trigger point value are needed to be learned from the training data. In the following sections, we first introduce the overall model learning procedure in Section.~\ref{sec:learning}, and then report our training details in Section.~\ref{sec:training}.

\subsection{Learning Procedure}
\label{sec:learning}


In the first stage, all parts are trained jointly over training set $S_{T}$ and validated over validation set $S_{V}$. In each epoch, the accuracy of all parts are evaluated over the validation. The model with the highest accuracy over the deepest part will be selected as the best model due to reaching the best possible accuracy at critical inference time.



\textbf{Identifying the trigger point:} Following the aforementioned design, the shallow part after feed-forwarding each input image has a confidence value over the output belief vector. To have an evaluation over this value and its range, we feed all images from $S_{T}$ into shallow part and collect the confidence values. The calculated mean $C_{Mean}$ and the standard deviation $C_{Std}$ over these values are used to control the expected classification accuracy of the AH-CNN. 

\subsection{Model Training Details}
\label{sec:training}

  
\textbf{Initializing:} We first adopt the ResNet-18 model as the base model, where each of the blocks in this model is considered as a separate classification module. We added a pooling and a fully connected layer for each part. Xaviar initialization \cite{glorot2010understanding} is used for having proper initial weights to propagate the signals precisely.    
    
\textbf{Defining the loss function:} For a classification task, the cross entropy is mostly used as loss function. Here, we have several parts which get their input from previous layer and have independent classification layer output. Consequently, these parts should be trained jointly. The objective function can be formulated as  
\begin{quote}
	\begin{center}
$L(\hat{y},y;\theta)=\sum_{N}L(\hat{y}_{n},y;\theta)$,
\end{center}
\end{quote}
where
\begin{quote}
	\begin{center}
		$L(\hat{y},y_{n};\theta)=-\frac{1}{\zeta_{S_{T}}}\sum_{k}y_{n}^{k}\log f(x_{k};\theta)$,
	\end{center}
\end{quote}
and $N$ denotes the total number of classification modules, $x_{k}$ the input images, $\zeta$ the set of all possible labels and $f(\theta)$ denotes the whole model.  

\begin{figure}[ht!]
    \centering
    \includegraphics[trim=6cm 0.8cm 5cm 0.8cm, clip=true, width=7.0cm]{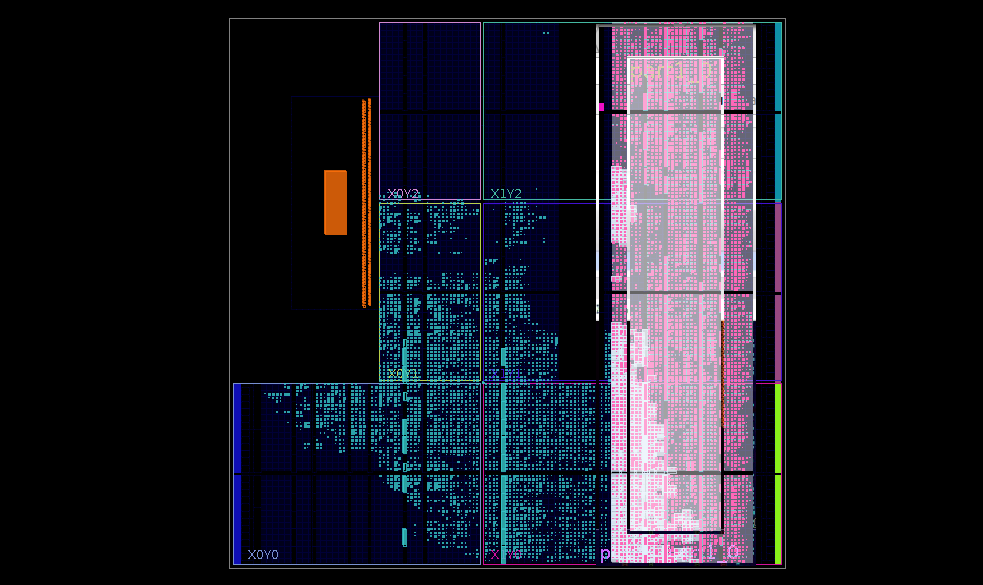}
    \caption{Layout of the reconfigurable design.}
    \label{layout}
\end{figure}

\section{Experiments} 
 

The theoretical framework we have presented suggests two hypotheses that deserve empirical tests: 1) AH-CNN can perform visual classification with much higher efficiency while maintaining the accuracy; and 2) Deep CNN models can be executed on a resource-constrained FPGA using partial reconfiguration. 
To validate these two hypotheses, we implement AH-CNN on Xilinx Zynq-7000 and evaluate  on the CIFAR-10, CIFAR-100 \cite{krizhevsky2009learning} and SVHN \cite{netzer2011reading}  datasets. We implemented the AH-CNN as described in Sec.~\ref{sec:method} where all convolution parts were implemented as separate hardware IP cores. We utilize Vivado HLS to synthesis the IP cores \cite{feist2012vivado}. The training procedure was  performed using the PyTorch framework. 

\subsection{Implementation}

We select the PYNQ-Z1 to perform our evaluations. This board consists of a Xilinx Zynq-7000 ZC7020 and a dual-core ARM A9 processor. Images were loaded to our convolution IP cores through a Direct Memory Access (DMA) IP core.    

We adopt the Resnet-18 \cite{he2016deep} as the base model. Due to limited available LUTs on this board, the network was broken into three parts. All parts consist of a group of convolution layers, pooling and fully connected layer. To reduce the reconfiguration time, we remove the last pooling and fully connected layer and create a new part which will be shared. Figure \ref{fig:CNNFPGA} shows an overview of our model. Part 1 is the shallowest model of this architecture. Parts 2-3 are the deeper blocks for extracting more features. Part 4 is the common one among all. Table \ref{resources} shows the resources needed for each part and total available resources on FPGA.

\begin{figure}[ht!]
	\begin{center}
		
		\includegraphics[trim=1.0cm 1cm 1.0cm 1cm,scale=0.4]{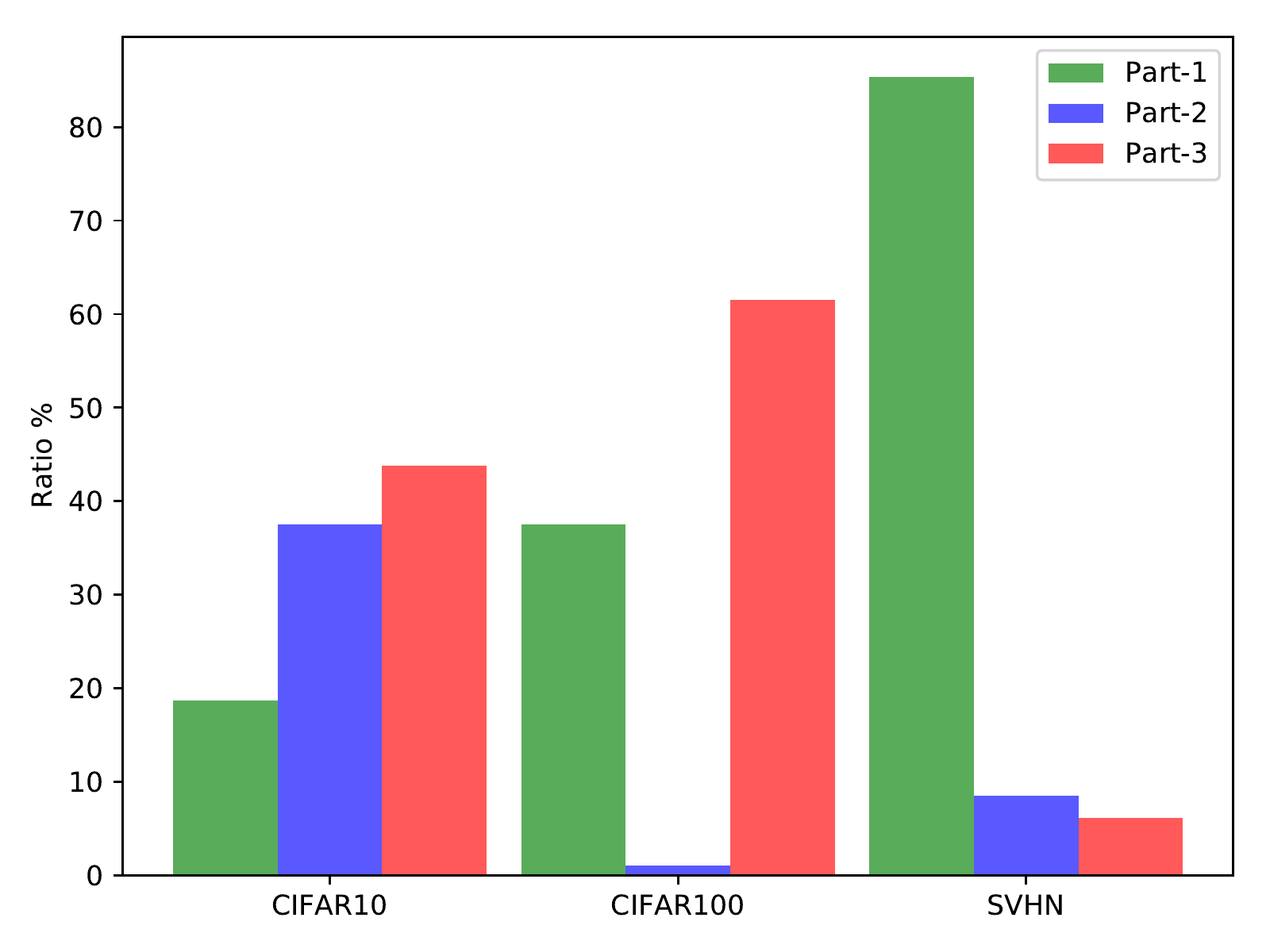}
	\end{center}
	\caption{The stop ratio for each part on CIFAR-10, CIFAR-100, and SVHN dataset.}
	\label{fig:Ratio}
\end{figure}
\begin{table}[h]
 \centering
\begin{tabular}{|c|c|c|c|c|c|}
\hline
              & \textbf{Part 1} & \textbf{Part 2} & \textbf{Part 3} & \textbf{Part 4} & \textbf{Total} \\ \hline
\textbf{BRAM} & 81              & 91              & 96              & 31              & 280            \\ \hline
\textbf{DSP}  & 120             & 96              & 96              & 24              & 220            \\ \hline
\textbf{FF}   & 15672           & 16647           & 34069           & 9908            & 106400         \\ \hline
\end{tabular}
\caption{Available resources on the Zynq XC7Z020, in comparison to used resources by convolution parts.}
\label{resources}
\end{table}

As shown previously, the total hardware resources needed for the whole architecture is more than available resource over the target device. Moreover, there are shared modules over all convolution parts such as Part 4, DMA, etc. Consequently, we applied Dynamic Partial Reconfiguration in order to reduce the reconfiguration time by just changing the convolution parts and keeping the shared modules. Fig.~\ref{layout} shows the layout of our implementation. The reconfigurable area is shown by purple and the fixed ports on the FPGA by white.

The resulting partial parts have all the same size of 2.4 MB and the size of main bitstream is 4 MB.

\textbf{Training:} The training part was carried out using PyTorch framework. We implemented special quantized convolution layer and fully connected layer with 1-bit weight and 5-bit activation.  The initial learning rate is set to be $0.01$ and it was decreased by a factor of $10$ in every $20$ epochs. Training continues until $100$ epochs with a mini-batch size of $256$.

\textbf{Feedback Evaluation:} The aforementioned procedure in section.~\ref{sec:training} is followed to estimate the confidence value. The mean and the standard deviation of all the confidence values were achieved after the various parts were collected over $S_{T}$.

\subsection{Overall Evaluation}
We choose the CIFAR10, CIFAR-100, and SVHN validation sets in the overall AH-CNN model testing. Here, we evaluate the partial reconfiguration approach. Also, we compare three selection methods: 1) our proposed feedback procedure; 2) SkipNet method \cite{wang2018skipnet};  and 3) an entropy-based method \cite{bolukbasi2017adaptive}.

\begin{figure*}[t!]
\centering
\begin{subfigure}{.33\textwidth}
  \centering
  \includegraphics[trim=0.0cm 0cm 0.0cm 0cm, clip=true, width=6.6cm]{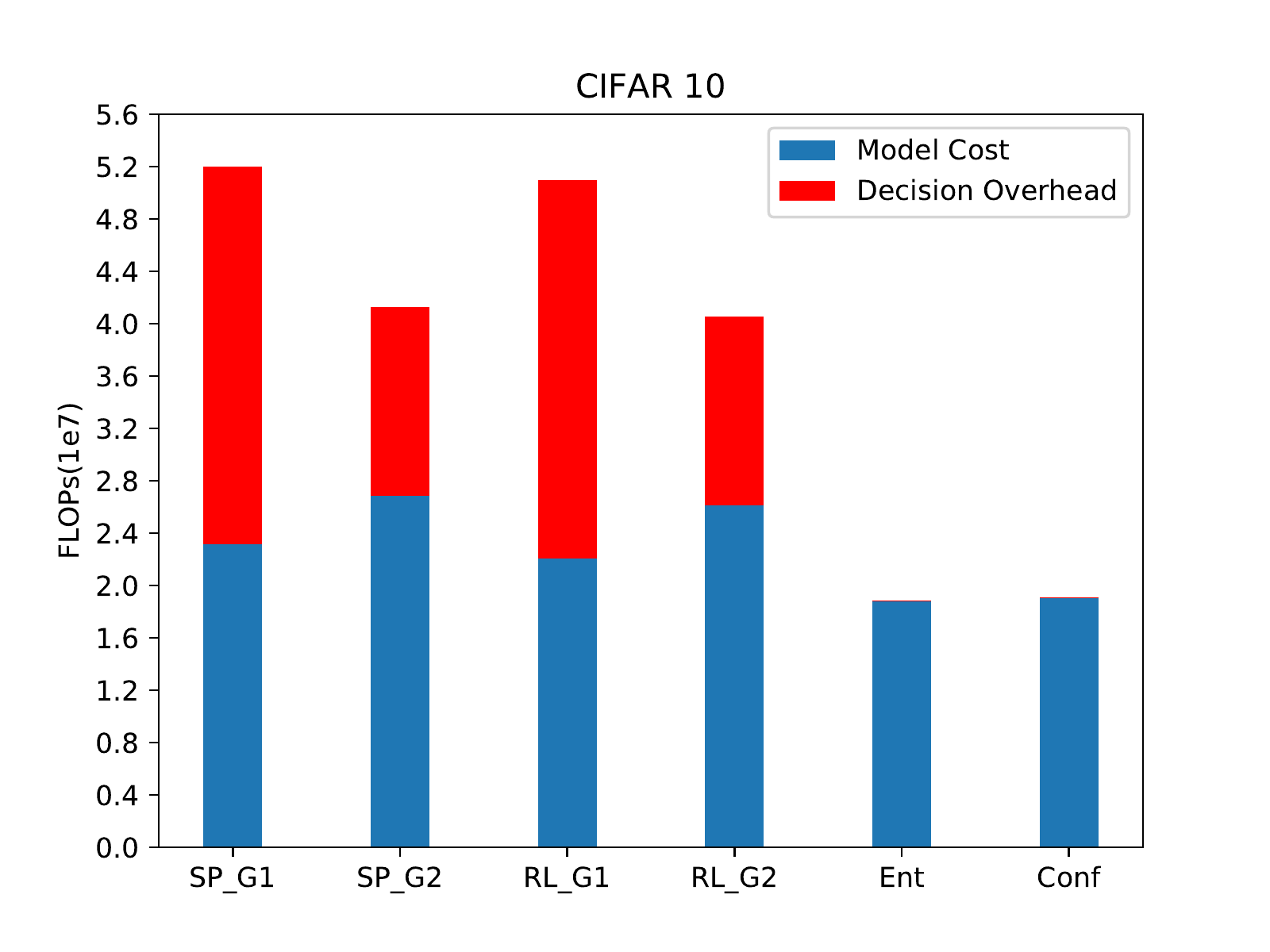}

  \label{DS:sub1}
\end{subfigure}%
\begin{subfigure}{.33\textwidth}
  \centering
  \includegraphics[trim=0cm 0cm 0.1cm 0cm, clip=true, width=6.6cm]{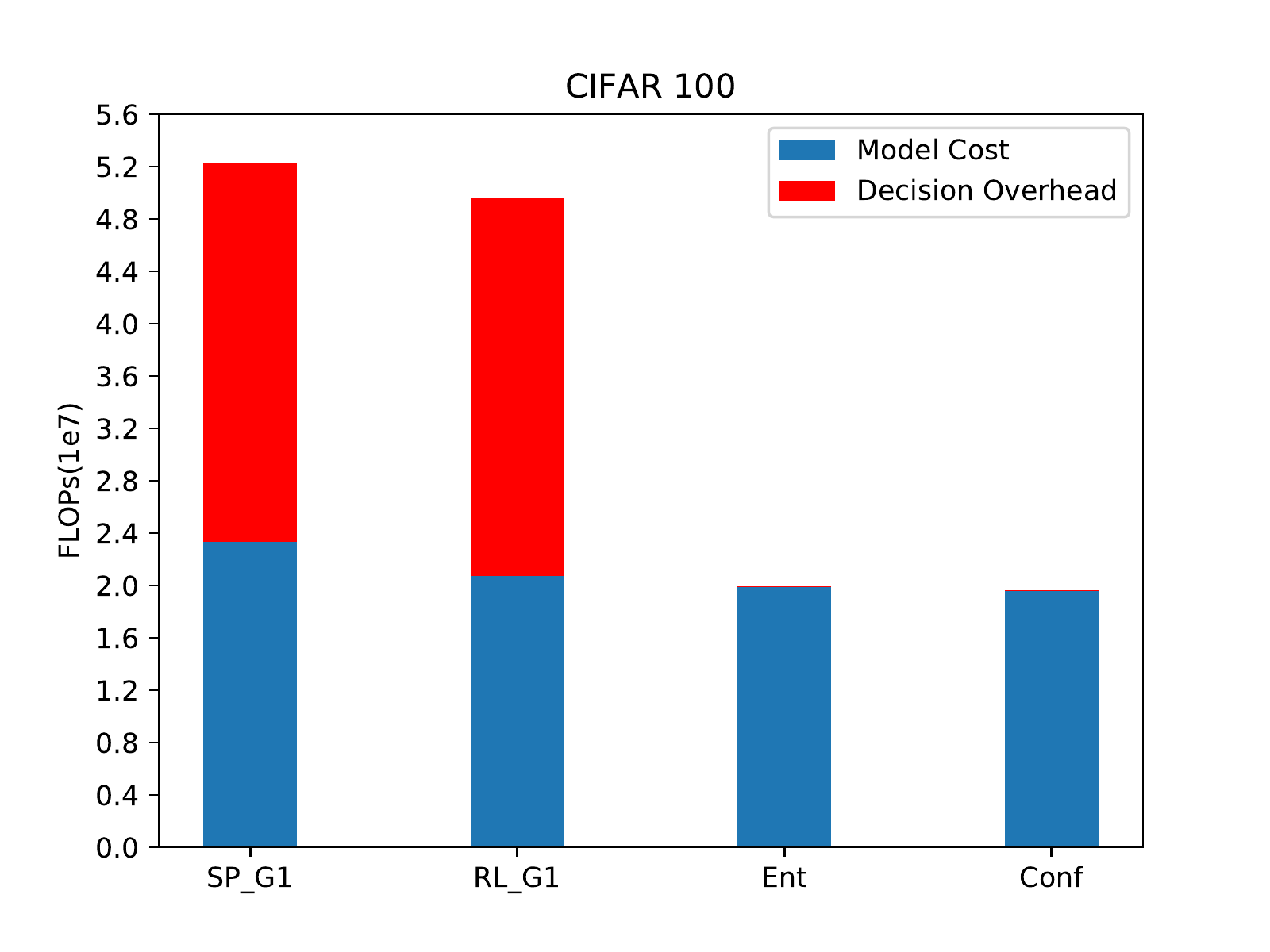}
 
  \label{DS:sub2}
\end{subfigure}
\begin{subfigure}{.334\textwidth}
  \centering
  \includegraphics[trim=0cm 0cm 0.1cm 0cm, clip=true, width=6.6cm]{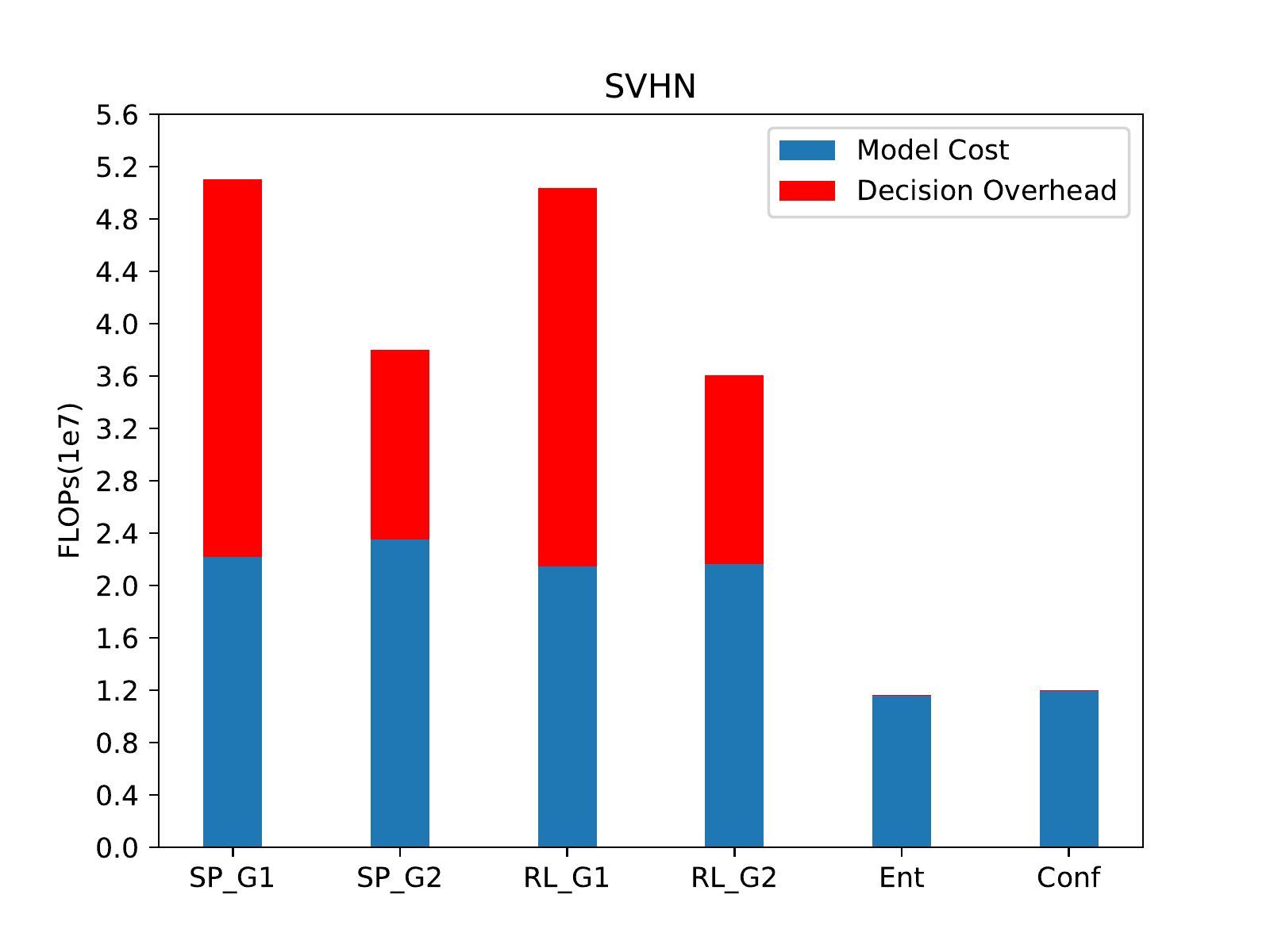}
 
  \label{DS:sub3}
\end{subfigure}
\caption{Computation reduction of Entropy (Ent), Confidence (Conf), SkipNet+SP (SP) and SkipNet+HRL+SP (RL) with feed-forward gates (G1 has two convolution layers, G2 has one convolution layer) while preserving the full network accuracy. The computation cost includes the computation of decision method. We are able to reduce the computation costs by $\approx$30\%, $\approx$27\% and $\approx$57\% on the CIFAR-10, CIFAR-100, and SVHN data using confidence decision method compared to the base model. Since the feed-forward gates are more expensive, SkipNet is not a suitable method for the scope of the study.}
\label{computation_compare}
\end{figure*}

\textbf{Partial Reconfiguration:} We have three accelerator IPs to reconfigure which are connected to the ARM processor through AXI interface, clocked at 100 MHz. The AXI channel and partial reconfiguration module is controlled by a Python script. We have also implemented a CPU version of AH-CNN architecture which runs on an ARM chip at 666 MHz. Table \ref{PR-TIME} shows the measurements of partial reconfiguration, FPGA and CPU execution time. As the reconfiguration region is same for all IPs, The reconfiguration time is always the same. By using batch processing (batch=512), the throughput of our system is $\approx$160 image per second while applying all parts to the images. This is 32 times faster than the CPU implementation.

\begin{table}[h]
 \centering
\begin{tabular}{|c|c|c|c|c|}
\hline
Bitstream & \begin{tabular}[c]{@{}c@{}}FPGA\\ Config\\ Time\end{tabular} & \begin{tabular}[c]{@{}c@{}}FPGA \\ Execution\\ Time\end{tabular} & \begin{tabular}[c]{@{}c@{}}CPU\\ Execution\\ Time\end{tabular} & \begin{tabular}[c]{@{}c@{}}FLOPS\end{tabular} \\ \hline
Part 1    & 38-42 ms                                                            & 2 ms                                                             & 98 ms                                                          & 10.24M                                                  \\ \hline
Part 2    & 38-42 ms                                                            & 2 ms                                                             & 57 ms                                                          & 8.6M                                                    \\ \hline
Part 3    & 38-42 ms                                                            & 2 ms                                                             & 49 ms                                                          & 8.5M                                                    \\ \hline
\end{tabular}{}

\caption{Performance evaluation on different parts of the design.}
\label{PR-TIME}
\end{table}

\begin{table}[h!]
 \centering
\begin{tabular}{c|c|c|c|c|}
\cline{2-5}
\multirow{2}{*}{}                     & \multirow{2}{*}{\textbf{CIFAR10}} & \multicolumn{2}{c|}{\textbf{CIFAR100}} & \multirow{2}{*}{\textbf{SVHN}} \\ \cline{3-4}
                                      &                                   & \textbf{Top1}      & \textbf{Top5}     &                                \\ \hline
\multicolumn{1}{|c|}{\textbf{Part 1}} & 70.95                             & 42.26              & 72.14             & 80.35                          \\ \hline
\multicolumn{1}{|c|}{\textbf{Part 2}} & 80.57                             & 52.23              & 80.25             & 91.24                          \\ \hline
\multicolumn{1}{|c|}{\textbf{Part 3}} & 86.27                             & 56.60              & 83.46             & 94.62                          \\ \hline
\end{tabular}
\caption{Top-1 accuracy of the HLS optimized IP-cores.}
\label{accuracy_PR}
\end{table}


Table \ref{accuracy_PR} shows the accuracy that can be achieved by applying each IP of convolutions to the input stream. It is clear that the system can reach to the higher accuracy by extracting more feature using deeper layers. Also, a significant portion of images can be classified correctly without using deep layers.

\noindent\textbf{Feedback Procedure:} Initially, we explore the trigger point by collecting the confidence of each AH-CNN branch. AH-CNN model achieves $85.4\%$, $55.4\%$, $94.2\%$ Top-1 validation accuracy over CIFAR10, CIFAR-100, and SVHN respectively. In Fig.~\ref{fig:Ratio}, we also report the portion of images classified by each branch. Due to the simplicity of the feedback procedure, this method has the lowest overhead.

\noindent\textbf{SkipNet \cite{wang2018skipnet}:} In this method, instead of selecting images by our feedback procedure, decision layer selects images using a gate consisting of convolution and fully connected layers. We adopt two different gates and two training methods proposed by \cite{wang2018skipnet} to evaluate our method. These gates show desirable performance over large CNN models. However they do not have the same performance over models such as ResNet-18 or ResNet-38. For each decision, one or two convolution layers and a fully connected layer should be applied to the stream. 

\noindent\textbf{Entropy Selection \cite{bolukbasi2017adaptive}:} This method uses the entropy of the shallow part's output to decide whether the input image needs further processing or not \cite{bolukbasi2017adaptive}. The work \cite{bolukbasi2017adaptive} implemented two variants: two-stacked model (AlexNet \cite{Krizhevsky2012} and -50 \cite{he2016deep} and  three-stacked model (AlexNet, GoogleNet \cite{szegedy2015going} and ResNet-50). Due to calculating the entropy of the output vector at each branch, this method is more expensive than the feedback procedure.

Fig.~\ref{computation_compare} depicts the computation reduction by applying the different decision procedures. We observe that by just considering the confidence, the model outperforms the SkipNet gates. SkipNet gates not only are so expensive but also are not as successful as other methods in our case study. The confidence and entropy selection have the same results however the confidence method has less computation cost. The confidence selection method decreased the computation to $69.8\%$, $71.8\%$, $43.8\%$ of the base model in CIFAR-10, CIFAR-100, and SVHN respectively. Also, the throughput of model reaches to $268$, $217$, and $408$ images per second.

\section{Conclusion}

In this paper, we proposed a new approach to run heavy neural networks on FPGAs with constrained resources. We stacked various shallow and deep models yielding an adaptive and hierarchical structure for quantaized neural networks. We conducted experiments on CIFAR-10, CIFAR-100 and SVHN, and empirically validated that AH-CNN maintains a similarly low inference time as the shallow models while achieving the high recognition accuracy of the deep model on image classification tasks. The flexible nature of this hierarchical method makes it suitable for applications that need adaptive behavior towards dynamic priority change over object categories, such as an agent with active perception.       

\noindent{\bf Acknowledgments: }The National Science Foundation under the Robust Intelligence Program (1750082), and the IoT Innovation (I-square) fund provided by ASU Fulton Schools of Engineering are gratefully acknowledged. We also acknowledge NVIDIA and Xilinx  for the donation of GPUs and FPGAs. 

{\tiny
\bibliographystyle{ieee}
\bibliography{egbib,AHNN}

}

\end{document}